\documentclass[12pt]{article}

\usepackage{sbc-template}
\usepackage{graphicx,url}
\usepackage[utf8]{inputenc}
\usepackage{placeins}
\usepackage{tabularx}
\usepackage{doi}
     
\title{New Product Development (NPD) through Social Media-based Analysis by Comparing Word2Vec and BERT Word Embeddings}

\author{Princessa Cintaqia\inst{1, 2}, Matheus Inoue\inst{2}}

\address{Department of Economics \\University of Indonesia -- Depok, West Java -- Indonesia
\nextinstitute
  Research and Development Team\\
  CloudWalk Inc -- São Paulo, SP -- Brazil
  \email{princessa.victory@ui.ac.id, matheus.inoue@cloudwalk.io}
}

\begin{document} 

\maketitle

\begin{abstract}
This study introduces novel methods for sentiment and opinion classification of tweets to support the New Product Development (NPD) process. Two popular word embedding techniques, Word2Vec and BERT, were evaluated as inputs for classic Machine Learning and Deep Learning algorithms to identify the best-performing approach in sentiment analysis and opinion detection with limited data. The results revealed that BERT word embeddings combined with Balanced Random Forest yielded the most accurate single model for both sentiment analysis and opinion detection on a use case. Additionally, the paper provides feedback for future product development performing word graph analysis of the tweets with same sentiment to highlight potential areas of improvement.
\end{abstract}
     
% \begin{resumo} 
%   Este meta-artigo descreve o estilo a ser usado na confecção de artigos e
%   resumos de artigos para publicação nos anais das conferências organizadas
%   pela SBC. É solicitada a escrita de resumo e abstract apenas para os artigos
%   escritos em português. Artigos em inglês deverão apresentar apenas abstract.
%   Nos dois casos, o autor deve tomar cuidado para que o resumo (e o abstract)
%   não ultrapassem 10 linhas cada, sendo que ambos devem estar na primeira
%   página do artigo.
% \end{resumo}

\section{Introduction}

New Product Development (NPD) can be defined as the process of developing new methods and features to improve the quality of current subsisting products \cite{sanayei2016technology}. NPD has often been projected as one of the key determining factors for the performance of firms \cite{zhan2021social}, however, NPD used to be a highly complex procedure \cite{ledwith2009market}. For example, customer feedback, which is sought to develop novel ideas, is retrieved through costly and inefficient surveys, focus groups, and product trials \cite{gozuacik2021social}, a process that is deemed too obtuse and slow for current fast-paced businesses in the digital era. 

Social media emerged as one of the most effective solutions to traditional customer feedback retrieval. Since NPD is partly dependent on customer involvement, product feedback retrieved from social media is a much cheaper and dynamic option \cite{saura2021using}. Therefore, social-media based sentiment analysis is often used to know whether customers like or dislike certain products \cite{saura2019three}. However, as written in \cite{gozuacik2021social}, sentiment analysis itself is not sufficient in giving desirable impact to the process of NPD. Although customer satisfaction, which is retrieved through sentiment analysis, is important, opinions from customers are needed to develop new features that are relevant to market demand. Hence, this study will present a method where both sentiment analysis and innovative opinions from customers can be analyzed through the detection of opinions from tweets about a launched product.

This study focus on doing sentiment analysis and opinion detection on social media data on a use case of US Airlines tweets. Sentiment analysis is done to see whether the context of a certain tweet is negative, positive, or neutral. Opinion detection is done to see whether a certain tweet has a valuable opinion for the product development process. To achieve this main objective, we compare Word2Vec \cite{mikolov2013efficient} and BERT \cite{devlin2018bert} word embeddings for both the sentiment analysis and the opinion detection tasks. A comparison between classic Machine Learning and Deep Learning algorithms is also done in this study to figure out which performs best with a small-to-medium amount of data.
Therefore, the objectives of this study are as follows: \begin{itemize}
\item Compare performances between word embeddings achieved through Word2Vec and BERT for both sentiment analysis and opinion detection tasks.
\item Compare performances between traditional Machine Learning and Deep Learning algorithms in handling small-to-medium amount of data.
\item Provide NPD suggestions for the product based on the analysis achieved through the sentiment analysis and opinion detection tasks on a use case.
\end{itemize}

\section{Related Work} \label{sec:headings}

NPD consists of four main phases: concept, development, validation and manufacturing \cite{sanayei2016technology}. All phases serve the main purpose of NPD: to develop existing products to suit the demand of the market. Social media is deemed as the easiest and cheapest customer-feedback tool for both big and small enterprises to understand what the current market demands, since, other than clients, social media data can also provide valuable information to a firm’s competitors, which is also one of the main variables to consider when proposing ideas during the development phase of the NPD. NPD itself does not only focus on adding specific features to suit market demands but also to execute some characteristics of the product which are found to be unpopular with customers.

The main inspiration used for this study is \cite{gozuacik2021social}. The authors believed that Deep Learning will enable them to extract both sentiments and innovative ideas for the NPD process from customers on social media (Twitter). They also expressed confidence that a combination of Deep Learning and advanced word embedding methods such as Word2Vec and GloVe \cite{pennington2014glove} maximizes feedback retrieval. It is important to note that in contrast to  \cite{gozuacik2021social}, in this study we also tested a non-Deep Learning algorithm and used embeddings from BERT \cite{devlin2018bert}.

A related work mainly comparing between Word2Vec and BERT is \cite{konstantinov2020approach} where they performed sentiment analysis with social media data. They used LSTM, a Deep Learning Neural Network architecture, to conduct the analysis. Results of the work shows that BERT performs better than Word2Vec. Another essential research concerning the comparison of Deep Learning and classic Machine Learning algorithms highlights the fact that Deep Learning itself has made doing heavy tasks possible \cite{chen2020deep}. The good results of Deep Learning, however, came together with expensive computational resources and time while also known to perform badly when trained under limited data \cite{brigato2021close}. This is why when served with a small-to-medium amount of data, classic Machine Learning algorithms are more often used than their deep learning counterpart \cite{kokol2022machine}.

\section{Methodology}

This study compares a classic word embedding technique, namely Word2Vec, to that of a transformer, BERT, with classic Machine Learning and Deep Learning algorithms. The goal of the comparison is mainly to see whether a change of the word embedding technique would increase the performance of the model and which method is best to use for the specific goal of sentiment analysis and opinion detection. Sentiment analysis is done to see whether the context of a certain tweet is negative, positive, or neutral. On the other hand, opinion detection is done to see whether a certain tweet has a valuable opinion for the product development process.

Afterwards, we compare the results of a classic Machine Learning algorithm to those of Deep Learning. The goal of this comparison is to see which of the two performs best with a small-to-medium amount of data used in this study. Finally, we will then use the best performing model to build graphs of words that will be able to help the NPD process by giving out recommendations achieved from the sentiments and opinions of the tweets. 

\subsection{Dataset}

The dataset being used in this study is extracted from Kaggle and is called the “Twitter US Airline Sentiment” dataset\footnote{https://www.kaggle.com/datasets/crowdflower/twitter-airline-sentiment} prepared by Crowdflower’s Data for Everyone library. The dataset consists of 14,640 English tweets, each one reviewing about one of the 6 airline brands present in the US. The Twitter US Airline Sentiment dataset was chosen as it already labels the sentiment of each tweet as either positive, neutral, or negative. Negative has the highest number of tweets with 9,178 (62.69\%), followed by neutral with 3,099 (21.16\%), and lastly positive with 2,363 (16.14\%). Each negative tweet is labeled with a reason on why the tweet is negative such as “Bad Flight”, “Late Flight”, etc, so this dataset can also be used for opinion detection. Table~\ref{tab:exTable1} shows a sample of the tweets and sentiments from the dataset.

\begin{table}[ht]
\small
\centering
\caption{Exemplary Tweets from the Original and Manually-labeled Dataset}
\label{tab:exTable1}
\begin{tabular}{c c c} 
\hline
 \textbf{Text} & \textbf{Sentiment} & \textbf{Opinion}\\
 \hline \\
When can I book my flight to Hawaii? & Neutral & No\\ \\
may start service to Hawaii from \#SanFrancisco this year & Neutral & Yes\\ \\
Very nicely done. & Positive & No\\ \\
it was amazing, and arrived an hour early. You're too good & Positive & Yes\\ \\
I will not be flying you again & Negative & No\\ \\
called your service line and was hung up on.\\ This is awesome sarcasm & Negative & Yes\\ \\
 \hline
\end{tabular}
\end{table}

\subsection{Data Preprocessing}
For sentiment analysis, we only extracted the “text” column of the dataset where it consists of the original tweets and the “airline sentiment” column where it gives the labeled positive, neutral, or negative sentiments of each tweet. We applied preprocessing operations to the “text” column, like removing problematic characters, stopwords were removed out using the a stopword list from Spacy \cite{honnibal2020spacy}. 
 
 For the opinion detection task, we manually labeled the first 3000 tweets from the dataset with a binary variable. Tweets that are labeled as "yes" mean that they contain valuable opinions for the NPD process. The authors of these tweets each pointed out a specific aspect they they liked or disliked about the product and gave solid reasons for their sentiments. In the contrary, tweets that are labeled as "no" mean that they simply do not contain any valuable opinions for the NPD process. The authors of these tweets tend to only show sentiments without showing any satisfying reason on why they feel that certain way towards the product. Tweets that are interrogative are also included in this "no" category. After the labelling process, we applied the same preprocessing techniques with the ones used in the sentiment analysis task. Table~\ref{tab:exTable1} presents the added opinion detection labels to the available tweets. 
 
 \subsection{Text Representation}
 Word embedding is an advanced text representation technique in Natural Language Processing (NLP) where words or phrases from the vocabulary are converted to vectors of real numbers. In this study, Word2Vec with transfer learning will be used. Word2Vec itself is a neural network-based model that translates words from a corpus into vectors with contextual comprehension \cite{mikolov2013efficient}. A pretrained Word2Vec model from a Google News dataset with 3 million vectors, 300 dimensions, and approximately 100 billion English words is used in this study.

BERT (Bidirectional Encoder Representations from Transformers) is a state-of-the-art language model that is able to deal with many NLP tasks such as sentiment analysis, question answering, and summarization \cite{devlin2018bert}. One example of a BERT application is the matter of subject-specific classification of texts \cite{yu2019improving}. In traditional methods or each word that is inserted into the embedding generation process would yield the same fixed vector, unconcerned about the real context of the word. But BERT, however, can highlight the polysemantic nature of words as it generates vectors according to the context by looking at the surrounding words. In this study, a lighter version of BERT, called DistilBERT is used as it is found to be 60\% faster, 40\% smaller, and on-device friendlier \cite{sanh2019distilbert}.

\subsection{Modelling}
The Gensim library \cite{vrehuuvrek2011gensim} is used to obtain the pretrained Word2Vec model which is the Google-news-300. For BERT, DistilBERT from Huggingface \cite{sanh2019distilbert} is used while also using Pytorch \cite{paszke2019pytorch} to get the word vectors. First of all, we split the dataset into training and test sets of 70\% and 30\%, respectively. We used 20\% of the training set as a validation set to find the most optimal hyperparameters. For hyper-parameter optimization, RandomizedSearchCV from Scikit Learn \cite{pedregosa2011scikit} is used. Balanced Random Forest Classifier \cite{chen2004using} from Imbalanced Learn \cite{lemaitre2017imbalanced} is the chosen Machine Learning algorithm in this study since for the specific case of this study the dataset is highly imbalanced. The results were then evaluated in terms of overall F1 scores and accuracy for each sentiment and opinion classification tasks. 

In the second step, we tested the exact same word embedding techniques under a Deep Learning algorithm. The architecture of the Deep Learning algorithms consists of a multi-task network, with the vectors achieved by the word embedding techniques as input layers, and two outputs: The sentiment analysis and opinion detection neurons. Each network has two hidden layers, each of them followed by Dropout and Batch Normalization layers. In Dropout, some fractions of the input will be dropped to make the model learns only generalized patterns rather than noise patterns \cite{srivastava2014dropout}. Batch Normalization is used to help speed up training and reduce computational resources as the models tend to need fewer epochs when Normalization is added \cite{ioffe2015batch}. Similar to the previous Balanced Random Forest experiment, we examined the results in terms of overall F1 scores and accuracy for each sentiment and opinion classification tasks.

\subsection{Word Graphs for NPD Recommendations}
After getting the model results, we used the best model to build word graphs that would show opinion-filled tweets for the NPD process. To achieve this, we applied the best model to the rest of the unlabeled tweets for opinion detection. Afterwards, we dropped the tweets that are labeled as not having any opinions so that only tweets that contain recommendations for the NPD process are taken and visualized. There are 7,719 opinionated tweets in total. We divided the tweets into two groups: positive and neutral in one group, negative in the other. We then constructed a co-occurence matrix using TF-IDF \cite{tfidf} to group words according to their cosine similarities. Finally, we made the word graphs based on the matrix using InfraNodus\cite{paranyushkin2019infranodus}, a text network visualization tool that is used to build graphs and represent the texts as a network.
 
\section{Results}\label{sec:figs}
\subsection{Modelling}
Model results were evaluated and ranked in terms of F1 scores. F1 score is a better evaluation metric than accuracy when the dataset is imbalanced such as the one used in this study. F1 score takes precision and recall into account, making it suitable for evaluating model performance when faced with the minority class of the data \cite{f1score}. 

As shown in Table~\ref{sentiment table}, for sentiment analysis, BERT is the better word embedding technique. Both BERT with Deep Learning and BERT with Balanced Random Forest beat the other two Word2Vec models. As for the comparison between Balanced Random Forest and Deep Learning, BERT with Deep Learning seems to surge as the winner against the other 3 models. With the highest F1 score (86.66\%), BERT with Deep Learning is able to classify positive, neutral, and negative tweets well even when the dataset is imbalanced and is small-to-medium sized. The two Balanced Random Forest models, one tested with BERT and the other with Word2Vec, seem to hold consistent results. Both Balanced Random Forest models yield rather similar results, even when paired with different word embedding techniques.

Meanwhile, the worst performing model for sentiment classification is Deep Learning combined with Word2Vec. This particular model has the highest accuracy, but the lowest F1 score out of all the other models. We found that Word2Vec with Deep Learning in the sentiment analysis task has a low precision score. This means that Word2Vec with Deep Learning experienced difficulties in handling the imbalance of the dataset labels. The model classified negative tweets as either positive or neutral to compensate for the lack of positive and neutral tweets. The high accuracy of the model only came from it classifying most of the negative tweets, the majority class, accurately.

\begin{table}[htp]
    \centering
    \caption{Sentiment Analysis Results}
    \begin{tabular}{cccc}
        \hline
        \textbf{Word Embedding} & \textbf{Model} & \textbf{F1 Score} & \textbf{Accuracy} \\ \hline
        BERT & Deep Learning & 86.66\% & 77.76\% \\
        BERT & Balanced Random Forest & 77.44\% & 76.41\%  \\
        Word2Vec & Balanced Random Forest & 75.32\% & 74.61\% \\
        Word2Vec & Deep Learning & 60.22\% & 91.10\% \\ \hline
    \end{tabular}
    \label{sentiment table}
\end{table}

\begin{table}[htp]
    \centering
    \caption{Opinion Detection Results}
    \begin{tabular}{cccc}
        \hline
        \textbf{Word Embedding} & \textbf{Model} & \textbf{F1 Score} & \textbf{Accuracy} \\ \hline
        BERT & Balanced Random Forest & 83.44\% & 81.19\% \\
        Word2Vec & Balanced Random Forest & 83.06\% & 80.51\%  \\
        Word2Vec & Deep Learning & 82.09\% & 80.06\% \\
        BERT & Deep Learning & 59.54\% & 84.56\% \\ \hline
    \end{tabular}
    \label{opinion table}
\end{table}

The opinion detection results are shown in Table~\ref{opinion table}. Although BERT with Balanced Random Forest is the best performing model for opinion detection, BERT with Deep Learning is the worst performing one. Therefore, a further analysis should be done to determine which word embedding technique is best in the case of detecting opinions. Thus, we should look at the comparison of the algorithm being used. Balanced Random Forest seems to be the model that performs best in the opinion detection task as both BERT with Balanced Random Forest and Word2Vec with Balanced Random Forest ranked higher than the two deep learning models. Again, with F1 scores that are rather similar, Balanced Random Forest seems to be consistent in its results regardless of the word embedding technique used.

Interestingly, BERT with Deep Learning, which was the best performing model for sentiment analysis, seems to be the worst one in terms of detecting opinions. Although having the highest accuracy, BERT with Deep Learning has low recall. This means that BERT with Deep Learning considers a lot of tweets with valuable opinions as not having any useful opinions for the NPD process. This is very alerting as one of the main objectives of the research is to find valuable opinions from customers to improve current and new products. If a model, such as BERT with Deep Learning, removes a lot of valuable opinions that are precious for product development, then only little, if any, improvement will be added to the products.

Looking at the results for sentiment and opinion classification simultaneously, we can say that BERT word embeddings is better than Word2Vec. However, the algorithm being used gives more weight in determining the performance of a certain model. In this case, Balanced Random Forest tends to be more consistent than Deep Learning. The results of Balanced Random Forest are stable even when combined with different word embedding techniques in both sentiment and opinion classification tasks. On the other hand, Deep Learning seems to be more inconsistent while needing more preprocessing steps. With the same text and data preprocessing pipeline, Balanced Random Forest seems to be able to handle the dataset better. The inconsistency of Deep Learning might also have to do with the fact that the dataset is small-to-medium sized. Therefore, we can say that BERT with Balanced Random Forest is the best single model to use if we want to do sentiment and opinion classification models simultaneously. 

\subsection{Word Graphs for NPD}
Both the positive and neutral, and also the negative tweets contained of mostly the same NPD recommendations. Both tweet groups are clustered into 4 clusters. We ordered the clusters according to importance of each as achieved through the "Main Topical Groups" feature in Infranodus. Here, the ranking of the clusters along with the most important words will be shown in percentage form. We can then click on any cluster to make it show in word graphs (visualization) form. The bigger a word is in the visualization, the more important is the word in the cluster.

\subsubsection{Cluster 1: Time Efficiency}
Time efficiency is the top factor customers of all of the US airlines consider important. Factors such as early arrival or departure, delayed and cancelled flights are the main things customers pay attention to, hence time efficiency being the most essential. Therefore, the main recommendation for all of the airlines in this case is to focus more on time efficiency matters such as reducing delays as much as possible. Figure 1 shows the most important words from the tweets included in Cluster 1.

\begin{figure}[ht]
\centering
\includegraphics[width=1\textwidth]{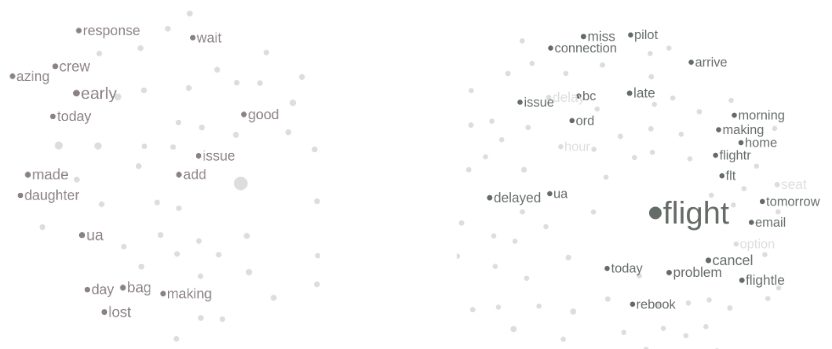}
\caption{Words from Positive and Neutral (left) and Negative (right) tweets about Time Efficiency}
\label{fig:Fig1}
\end{figure}
\FloatBarrier

\subsubsection{Cluster 2: Flight and Booking}
Flight and booking matters are the second most important factors to all of the airlines' customers. In tweets where sentiments are positive and neutral, customers praised how flights are still continued even under harsh weather conditions. In tweets where sentiments are negative, customers mostly complained about how they had to wait in long lines to check-in while also highlighting the complicated rebooking process. Figure 2 shows the most important words from the tweets included in Cluster 2.

\begin{figure}[ht]
\centering
\includegraphics[width=1\textwidth]{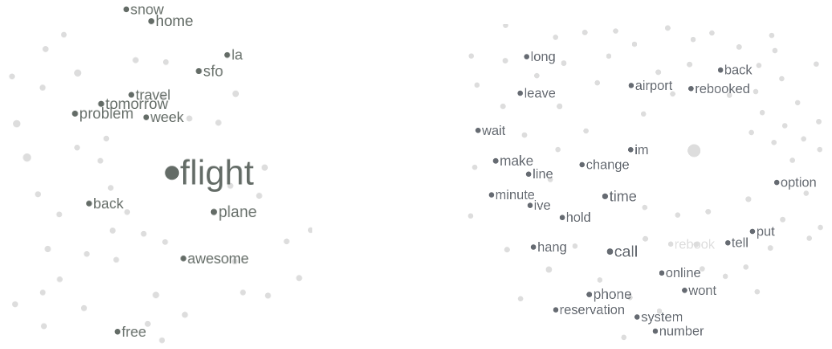}
\caption{Words from Positive and Neutral (left) and Negative (right) tweets about Flight and Booking}
\label{fig:Fig2}
\end{figure}
\FloatBarrier

\subsubsection{Cluster 3: Facilities}
The third most important factor for customers is facilities. This can go from good on-flight food, opening up rare routes around the country, comfortable seats, to the lack of on-flight wi-fi and movies for longer duration flights. Figure 3 shows the most important words from the tweets included in Cluster 3.

\begin{figure}[ht]
\centering
\includegraphics[width=1\textwidth]{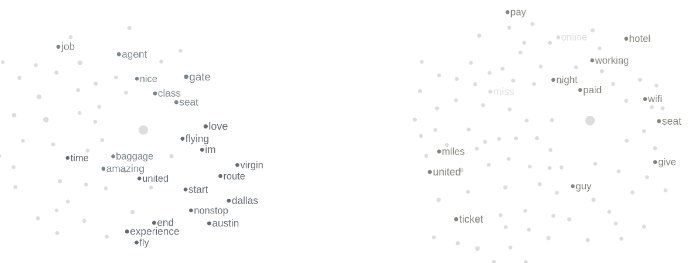}
\caption{Words from Positive and Neutral (left) and Negative (right) tweets about Facilities}
\label{fig:Fig3}
\end{figure}
\FloatBarrier

\subsubsection{Cluster 4: Customer Service}
Surprisingly, customer service placed last on the factors that customers consider as important. Positive tweets mostly commented on the helpfulness of the airlines' customer service, some explaining how they are able to get refund for cancelled flights easily. On the other hand, negative tweets commented on how unhelpful is the customer service mostly for not answering calls and messages. Figure 4 shows the most important words from the tweets included in Cluster 4.

\begin{figure}[ht]
\centering
\includegraphics[width=1\textwidth]{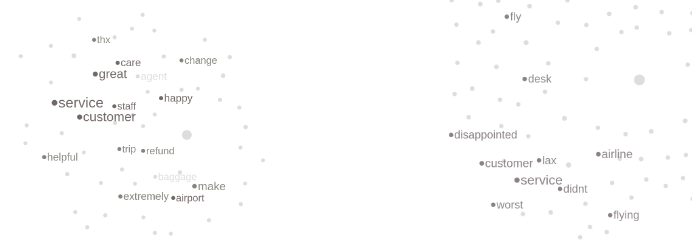}
\caption{Words from Positive and Neutral (left) and Negative (right) tweets about Customer Service}
\label{fig:Fig4}
\end{figure}
\FloatBarrier

\section{Conclusion}

In conclusion, our study revealed that models utilizing BERT word embeddings outperformed Word2Vec in both sentiment analysis and opinion detection tasks. Deep Learning demonstrated superior performance over Balanced Random Forest for sentiment analysis, while Balanced Random Forest models outperformed Deep Learning for opinion detection. We identified BERT with Balanced Random Forest as the best-performing single model, as it achieved competitive performance with reduced computational resources and consistent results even with limited data.

We also leveraged word graphs to provide valuable recommendations for New Product Development (NPD) in the airline industry. Our analysis indicated that the most important factors for customers, in order of priority, were time efficiency, flight and booking, facilities, and customer service. Therefore, addressing time efficiency issues, such as flight delays and cancellations, should be a primary focus for airlines to improve customer satisfaction.

Furthermore, our study contributes to the literature by proposing a faster and cost-effective method for retrieving recommendations from social media to support the NPD process. The methodology used in this study can be adopted by small-to-medium enterprises (SMEs) with limited data to improve their products using their own social media data. Future research can explore the use of larger datasets to further evaluate the performance of classic Machine Learning algorithms compared to Deep Learning in retrieving sentiment and valuable opinions for NPD.

\bibliographystyle{sbc}
\bibliography{sbc-template}

\begin{thebibliography}{}

\bibitem[Brigato and Iocchi 2021]{brigato2021close}
Brigato, L. and Iocchi, L. (2021).
\newblock A close look at deep learning with small data.
\newblock In {\em 2020 25th International Conference on Pattern Recognition
  (ICPR)}, pages 2490--2497. IEEE.

\bibitem[Chen et~al. 2004]{chen2004using}
Chen, C., Liaw, A., Breiman, L., et~al. (2004).
\newblock Using random forest to learn imbalanced data.
\newblock {\em University of California, Berkeley}, 110(1-12):24.

\bibitem[Chen et~al. 2020]{chen2020deep}
Chen, C., Zhang, P., Zhang, H., Dai, J., Yi, Y., Zhang, H., and Zhang, Y.
  (2020).
\newblock Deep learning on computational-resource-limited platforms: a survey.
\newblock {\em Mobile Information Systems}, 2020.

\bibitem[Devlin et~al. 2018]{devlin2018bert}
Devlin, J., Chang, M.-W., Lee, K., and Toutanova, K. (2018).
\newblock Bert: Pre-training of deep bidirectional transformers for language
  understanding.
\newblock {\em arXiv preprint arXiv:1810.04805}.

\bibitem[Gozuacik et~al. 2021]{gozuacik2021social}
Gozuacik, N., Sakar, C.~O., and Ozcan, S. (2021).
\newblock Social media-based opinion retrieval for product analysis using
  multi-task deep neural networks.
\newblock {\em Expert Systems with Applications}, 183:115388.

\bibitem[Honnibal et~al. 2020]{honnibal2020spacy}
Honnibal, M., Montani, I., Van~Landeghem, S., and Boyd, A. (2020).
\newblock spacy: Industrial-strength natural language processing in python.
\newblock To appear.

\bibitem[Ioffe and Szegedy 2015]{ioffe2015batch}
Ioffe, S. and Szegedy, C. (2015).
\newblock Batch normalization: Accelerating deep network training by reducing
  internal covariate shift.
\newblock In {\em International conference on machine learning}, pages
  448--456. PMLR.

\bibitem[Kokol et~al. 2022]{kokol2022machine}
Kokol, P., Kokol, M., and Zagoranski, S. (2022).
\newblock Machine learning on small size samples: A synthetic knowledge
  synthesis.
\newblock {\em Science Progress}, 105(1):00368504211029777.

\bibitem[Konstantinov et~al. 2020]{konstantinov2020approach}
Konstantinov, A., Moshkin, V., and Yarushkina, N. (2020).
\newblock Approach to the use of language models bert and word2vec in sentiment
  analysis of social network texts.
\newblock In {\em International Scientific and Practical Conference in Control
  Engineering and Decision Making}, pages 462--473. Springer.

\bibitem[Ledwith and O'Dwyer 2009]{ledwith2009market}
Ledwith, A. and O'Dwyer, M. (2009).
\newblock Market orientation, npd performance, and organizational performance
  in small firms.
\newblock {\em Journal of Product Innovation Management}, 26(6):652--661.

\bibitem[Lema{\^\i}tre et~al. 2017]{lemaitre2017imbalanced}
Lema{\^\i}tre, G., Nogueira, F., and Aridas, C.~K. (2017).
\newblock Imbalanced-learn: A python toolbox to tackle the curse of imbalanced
  datasets in machine learning.
\newblock {\em The Journal of Machine Learning Research}, 18(1):559--563.

\bibitem[Mikolov et~al. 2013]{mikolov2013efficient}
Mikolov, T., Chen, K., Corrado, G., and Dean, J. (2013).
\newblock Efficient estimation of word representations in vector space.
\newblock {\em arXiv preprint arXiv:1301.3781}.

\bibitem[Paranyushkin 2019]{paranyushkin2019infranodus}
Paranyushkin, D. (2019).
\newblock Infranodus: Generating insight using text network analysis.
\newblock In {\em The world wide web conference}, pages 3584--3589.

\bibitem[Paszke et~al. 2019]{paszke2019pytorch}
Paszke, A., Gross, S., Massa, F., Lerer, A., Bradbury, J., Chanan, G., Killeen,
  T., Lin, Z., Gimelshein, N., Antiga, L., et~al. (2019).
\newblock Pytorch: An imperative style, high-performance deep learning library.
\newblock {\em Advances in neural information processing systems}, 32.

\bibitem[Pedregosa et~al. 2011]{pedregosa2011scikit}
Pedregosa, F., Varoquaux, G., Gramfort, A., Michel, V., Thirion, B., Grisel,
  O., Blondel, M., Prettenhofer, P., Weiss, R., Dubourg, V., et~al. (2011).
\newblock Scikit-learn: Machine learning in python.
\newblock {\em the Journal of machine Learning research}, 12:2825--2830.

\bibitem[Pennington et~al. 2014]{pennington2014glove}
Pennington, J., Socher, R., and Manning, C.~D. (2014).
\newblock Glove: Global vectors for word representation.
\newblock In {\em Proceedings of the 2014 conference on empirical methods in
  natural language processing (EMNLP)}, pages 1532--1543.

\bibitem[Sammut and Webb 2010]{tfidf}
Sammut, C. and Webb, G.~I., editors (2010).
\newblock {\em TF--IDF}.
\newblock Springer US, Boston, MA.

\bibitem[Sanayei 2016]{sanayei2016technology}
Sanayei, A. (2016).
\newblock {\em Technology decisions in new product development}.
\newblock Wayne State University.

\bibitem[Sanh et~al. 2019]{sanh2019distilbert}
Sanh, V., Debut, L., Chaumond, J., and Wolf, T. (2019).
\newblock Distilbert, a distilled version of bert: smaller, faster, cheaper and
  lighter.
\newblock {\em arXiv preprint arXiv:1910.01108}.

\bibitem[Saura 2021]{saura2021using}
Saura, J.~R. (2021).
\newblock Using data sciences in digital marketing: Framework, methods, and
  performance metrics.
\newblock {\em Journal of Innovation \& Knowledge}, 6(2):92--102.

\bibitem[Saura and Bennett 2019]{saura2019three}
Saura, J.~R. and Bennett, D.~R. (2019).
\newblock A three-stage method for data text mining: Using ugc in business
  intelligence analysis.
\newblock {\em Symmetry}, 11(4):519.

\bibitem[Srivastava et~al. 2014]{srivastava2014dropout}
Srivastava, N., Hinton, G., Krizhevsky, A., Sutskever, I., and Salakhutdinov,
  R. (2014).
\newblock Dropout: a simple way to prevent neural networks from overfitting.
\newblock {\em The journal of machine learning research}, 15(1):1929--1958.

\bibitem[Umer et~al. 2020]{f1score}
Umer, M., Imtiaz, Z., Ullah, D.~S., Mehmood, A., Choi, G.~S., and On, B.-W.
  (2020).
\newblock Fake news stance detection using deep learning architecture
  (cnn-lstm).
\newblock {\em IEEE Access}, PP:1--1.

\bibitem[Yu et~al. 2019]{yu2019improving}
Yu, S., Su, J., and Luo, D. (2019).
\newblock Improving bert-based text classification with auxiliary sentence and
  domain knowledge.
\newblock {\em IEEE Access}, 7:176600--176612.

\bibitem[Zhan et~al. 2021]{zhan2021social}
Zhan, Y., Han, R., Tse, M., Ali, M.~H., and Hu, J. (2021).
\newblock A social media analytic framework for improving operations and
  service management: A study of the retail pharmacy industry.
\newblock {\em Technological Forecasting and Social Change}, 163:120504.

\bibitem[Řehůřek et~al. 2011]{vrehuuvrek2011gensim}
Řehůřek, R., Sojka, P., et~al. (2011).
\newblock Gensim—statistical semantics in python.
\newblock {\em Retrieved from genism. org}.

\end{thebibliography}


\begin{thebibliography}{}

\bibitem[Boulic and Renault 1991]{boulic:91}
Boulic, R. and Renault, O. (1991).
\newblock 3d hierarchies for animation.
\newblock In Magnenat-Thalmann, N. and Thalmann, D., editors, {\em New Trends
  in Animation and Visualization}. John Wiley {\&} Sons ltd.

\bibitem[Knuth 1984]{knuth:84}
Knuth, D.~E. (1984).
\newblock {\em The {\TeX} Book}.
\newblock Addison-Wesley, 15th edition.

\bibitem[Smith and Jones 1999]{smith:99}
Smith, A. and Jones, B. (1999).
\newblock On the complexity of computing.
\newblock In Smith-Jones, A.~B., editor, {\em Advances in Computer Science},
  pages 555--566. Publishing Press.

\end{thebibliography}

\end{document}